\documentclass[a4paper]{esannV2}
\usepackage[latin1]{inputenc}
\usepackage{amssymb,array}
\usepackage[fleqn]{amsmath}
\usepackage[pdftex]{graphicx}
\usepackage{hyperref}

\usepackage{ntheorem}

\makeatletter
\theoremseparator{.}
\newtheoremstyle{magyartetelstilus}%
{\item[\hskip\labelsep\theorem@headerfont
            ##2.\ ##1\theorem@separator]}%
{\item[\hskip\labelsep\theorem@headerfont
            ##2.\ ##1\theorem@separator\ (##3)]}%
\theoremstyle{magyartetelstilus}
\newtheorem{tetel}{Theorem}[section]

\newtheorem{defi}[tetel]{Definition}
{\theorembodyfont{\rmfamily}

}%

\makeatother

\begin{document}
\title{Gradient representations in ReLU networks as similarity functions\footnote{Accepted at the 29th European Symposium on Artificial Neural Networks, Computational Intelligence and Machine Learning (ESANN 2021), 6-8 October 2021. Source code with additional experiments: \url{https://github.com/danielracz/gradsim}}}

\author{D\'aniel R\'acz$^{1,3}$ and B\'alint Dar\'oczy$^{2,3}$\\
\thanks{BD was supported by MIS ``Learning from Pairwise Comparisons'' of the F.R.S.-FNRS and by MTA Premium Postdoctoral Grant 2018.}
\vspace{.3cm}\\
1 - E\"otvos Lor\'and University, Institute of Mathematics \\
P\'azm\'any P\'eter s\'et\'any 1/C, H-1117, Budapest, Hungary \\
2 - Universit\'e catholique de Louvain, INMA/ICTEAM\\
Avenue Georges Lema\^itre 4, B-1348, Louvain-la-Neuve, Belgium\\
3 - Institute for Computer Science and Control, SZTAKI/ELKH \\
Kende utca 13-17, H-1111, Budapest, Hungary \\
e-mail: racz.daniel@sztaki.hu, balint.daroczy@uclouvain.be}

\maketitle 

\begin{abstract}
Feed-forward networks can be interpreted as mappings with linear decision surfaces at the level of the last layer. We investigate how the tangent space of the network can be exploited to refine the decision in case of ReLU (Rectified Linear Unit) activations. We show that a simple Riemannian metric parametrized on the parameters of the network forms a similarity function at least as good as the original network and we suggest a sparse metric to increase the similarity gap. 
\end{abstract}

\section{Introduction}

We consider feed-forward neural networks with ReLU activations to examine how the network's final representation space connected to the gradient structure of the network. Our motivation is twofold: recently discovered knowledge about ReLU networks \cite{hanin2019complexity,zhang2020empirical,hanin2021deep} and recent results about higher order optimization methods \cite{martens2015optimizing}. In a way, many of the existing machine learning problems can be investigated as statistical learning problems, therefore information geometry \cite{amari1996neural} plays an important role. It was shown in \cite{ollivier2015riemannian} that over the parameter space of a neural network we can often determine a Riemannian manifold based on an error or loss function, moreover the tangent bundle on specific Riemannian metrics, e.g. Fisher information, has unique invariance properties \cite{Cencov1982,perronnin2007fisher}. 

Our main hypothesis is that in feed-forward ReLU networks we can utilize the relation of the parameters and the output due the homogeneity property of the activation functions therefore we investigate the gradient structure of the network output w.r.t. the parameters and exploit the space induced by the partial derivatives together with a metric as a representation of data points. Albeit the inner product space of the tangent bundle is quadratic, there are well-defined underlying structures in the tangent space specific to ReLU networks. In this paper we introduce several similarity functions \cite{balcan2008theory} based on a block-diagonal, sparse metric and we inspect how they relate to the similarity induced by the network itself by measuring the \textit{similarity gap}, the difference in expected similarity between point pairs with the same label and point pairs with different labels. 
 
\subsection{Related work}

The geometrical properties of the underlying loss manifold of neural networks was used as a general framework for optimization in classification \cite{ollivier2015riemannian} and for generative models \cite{rifai2011manifold}. Furthermore Martens and Grosse \cite{martens2015optimizing} approximated Amari's natural gradient \cite{amari1996neural} for feed-forward neural networks with block partitioning the Fisher information matrix by exploiting the structure of the network. The partial gradients were used as representations in classification given a known generative probability density function, e.g. through an approximated Fisher kernel, in \cite{jaakkola1999exploiting} where the authors approximated the metric with the diagonal of the Fisher information matrix, closed formula for Gaussian mixtures were proposed in \cite{perronnin2007fisher}. In \cite{daroczy2019tangent} the authors learned a metric with a neural network over the partial gradients w.r.t. the parameters. Not only the partial gradient w.r.t. parameters carry important properties e.g. in \cite{novak2018sensitivity} the authors experimented with the norm of the input-output sensitivity, the Frobenius norm of the Jacobian matrix of the output w.r.t. input in case of simple architectures. 


Recent results show that under simple presumptions the maximal ``capacity'' (representational power) of deep ReLU networks is related to the arrangement of the polytopes in the input space \cite{hanin2019complexity} and to the properties of transition between linear regions \cite{raghu2017expressive,hanin2021deep} instead of the exact number of polytopes with non-zero volume. 

Our last ingredient is Balcan and Blum's theory of similarity functions \cite{balcan2008theory}, which we will use as our foundation to show that we can define similarity functions in the gradient space that are at least as good similarity functions as the output of the network. 

\subsection{Notations}

Let $f: \mathbb{R}^{d} \times \mathbb{R}^{|\theta|} \rightarrow \mathbb{R}$ be a feed-forward ReLU network, as a function of both the input data and the weights, containing $L$ hidden layers, $N$ neurons trained for solving a classification problem $P$ for some data distribution $D$. The activation function is ReLU: $h(x) = \max(0, x), x \in \mathbb{R}$. Let $N^k$ denote the number of neurons in the $k$-th layer and let $h^k_{\theta}(x) \in \mathbb{R}^{N^k}$ denote the output of the $k$-th layer for a given input vector $x$ and a fixed set of weights $\theta$. Denoting the weight vector after the last hidden layer, the discriminative layer, by $\theta_L$ we have $f_{\theta}(x) = \theta_{L}^{T} h^L_{\theta}(x)$. We denote the set of parameters as $\theta$ and the parameters of the $k$-th layer with $\theta_k$. The actual output of our model is $sgn(f_{\theta}(x)) = sgn(\theta_{L}^{T} h_{\theta}(x))$, i.e the network can be interpreted as a linear separator acting on $Im(h_{\theta}) \subseteq \mathbb{R}^{N^L}$, the image of the nonlinear mapping $h_{\theta}: \mathbb{R}^{d} \rightarrow \mathbb{R}^{N^L}$. Following \cite{hanin2019complexity}, we define an \textit{activation pattern} with assigning a sign to each neuron in the network, $A = \{a_{l};l=1,..,N\} \in \{-1,1\}^N$. For a particular input we refer $A(x;\theta) = \{sign([h_{\theta}(x)]_l; l=1,..,N\}$ as the activation pattern assigned to an input $x$. An \textit{activation region} with the corresponding fixed $\theta$ and $A$ is defined as $R(A;\theta):=\{x \in \mathbb{R}^{d} | sign([h_{\theta}(x)]_{l}) = a_{l}\}$, the set of input assigned to the same activation pattern. In comparison, linear regions are the input regions where the function defines different linear regions.We define tangent vectors as the change in the output with a directional derivative of $f_{\theta}(x)$ in the direction of $g_{\theta}(x_i) = \frac{\partial f_{\theta}(x)}{\partial \theta} \big|_{x=x_i}$: $(D_{g_{\theta}(x_i)}f)(\theta)=\frac{d}{dt}[f_{\theta}(x)+tg_{\theta}(x_i))]|_{t=0}$. We will refer $\nabla: \mathbb{R}^{d} \rightarrow \mathbb{R}^{|\theta|}$ as the tangent mapping of input at $\theta$: $\nabla_\theta f_{\theta}(x) := \frac{\partial{f_{\theta}(x)}}{\partial {\theta}}$. We denote the partial gradient vectors with $g_{\theta}(x) \in \mathbb{R}^{|\theta|}$.

\section{Similarity functions}

First, we state the following definition from \cite{balcan2008theory} which will be the measure of our similarity functions: 

\begin{defi}
A $K(x,y)$ is a weakly $\gamma$-good similarity function for a learning problem P if:
\begin{equation}
E_{x,y \sim P}[K(x,y) | l(x) = l(y)] \geq E_{x,y \sim P}[K(x,y) | l(x) \neq l(y)]  + \gamma.
\end{equation}
\end{defi}

Note, the strong version of good similarity function says that a $K(x,y)$ is a good similarity function if $1-\epsilon$ probability mass of the examples are on average more similar to examples of the same category than to examples of another category. We consider the weakly good similarity because in this case we may investigate how the gap changes in case of a finite sample. We refer to $\gamma$ as the \textit{similarity gap}. 

We define the similarity function of the discriminative layer as $K_{f^L_{\theta}}(x,y) = h_{\theta}^L (x)^T h_{\theta}^{L}(y)$ and the similarity function of the network as $K_{f_{\theta}}(x,y) = f_{\theta} (x) f_{\theta}(y)$. Now, let us consider the inner product in case of the tangent mapping. Before we argue about the metric in the tangent space we take a closer look on the structure of the Hessian in case of feed-forward ReLU networks. By definition the Hessian $H_{\theta}$ of the function induced by the network for a single output is a $|\theta| \times |\theta|$ sized matrix with $[H_{\theta}(x)]_{ij} = \frac{\partial^2 [f_{\theta}(x)]}{\partial \theta_i \partial \theta_j}, \forall i,j \in \{1,..,|\theta|\}$ which in our case is equal to the sum over the paths from the input to the output node where both $\theta_i$ and $\theta_j$ are present therefore if we split the Hessian into sub-blocks, where only elements from the same layer are present, these sub-blocks will be diagonal. This property suggest us to seek for a metric where the parameters of a layer in these sub-blocks may collide if necessary. Therefore we suggest a simple metric $M_{\theta} \in \mathbb{R}^{|\theta|^2}$ with $[M_{\theta}]_{ij} = \delta_{l(i),l(j)} \theta_i \theta_j$ where $\delta_{i,j}$ is one if $i = j$ and zero otherwise and $l(i)$ is the index of the layer of the $i$-th parameter. We define our new similarity function, the \textit{block diagonal similarity} as
\begin{align*}
K_{M_{\theta}}(x,y) = g_{\theta}(x)^T M_{\theta} g_{\theta}(y) = \sum_{i,j \in \theta \times \theta} \delta_{l(i),l(j)} \theta_i \theta_j [g_{\theta}(x)]_i [g_{\theta}(y)]_j
\end{align*}
\noindent where we utilized that the metric has a Cholesky decomposition as it is positive semi-definite. One of the most important properties of the block-diagonal similarity function is that in case of a single output the similarity function is equal, up to a constant, to the network output as 
\begin{align*}
& \sum_{i,j \in \theta \times \theta} \delta_{l(i),l(j)} \theta_i \theta_j [g_{\theta}(x)]_i [g_{\theta}(y)]_j  = \sum_{i,j \in \theta \times \theta}  \delta_{l(i),l(j)}  f_{\theta}(x) f_{\theta}(y) \\
&=  \sum_{k=1}^L K_{f_{\theta}(x,y)} \approx O(K_{f_{\theta}}(x,y)).
\end{align*}
\noindent Since $K_{f_{\theta}}(x,y)  = \sum_{i \in \theta^L} \theta^2_i [h_{\theta}(x)]_i [h_{\theta}(y)]_i$ the following holds for the block-diagonal similarity: 
\begin{align*}
\omega_L O(K_{f^L_{\theta}}(x,y)) \leq K_{M_{\theta}}(x,y) \leq \omega_L^* O(K_{f^L_{\theta}}(x,y)) 
\end{align*}
where $\omega_L = \min_{i \in \theta^L} \theta_i^2$ and $\omega_L^* = \max_{i \in \theta^L} \theta_i^2$. Note, the similarity values are not necessary positive. 

\subsection{Sparse metric}

Now, we decompose the gap and suggest a modified metric to increase the gap of the block-diagonal similarity function. As the norm of the gradient vectors highly affects the gap, we normalize to avoid linear increases in the gap. First, for simplicity we estimate the expectation $\mathbf{E}_{x,y \sim P}[K_{M_{\theta}}(x,x') | l(x) = l(y)]$ for a single layer and for a single label therefore the expectation is 
\begin{align*}
&\sum_{i=1}^{|\theta|} \sum_{j=1}^{|\theta|}  \mathbf{E}_{x,x' \sim P}[\frac{[g_{\theta}(x)]_i \theta_i \theta_j [g_{\theta}(x')]_j}{\|M_{\theta}^{1/2} g_{\theta}(x) \| \| M_{\theta}^{1/2} g_{\theta}(x') \|}]  \\
&\approx \frac{1}{|T^{(+)}|^2} \sum_{t_1,t_2 \in T^{(+)} \times T^{(+)}} \frac{1}{\|\hat{g_{\theta}}(x_{t_1})\| \|\hat{g}_{\theta}(x_{t_1})\| }  \sum_{i=1}^{|\theta|} \sum_{j=1}^{|\theta|}  [g_{\theta}(x_{t_1})]_i \theta_i \theta_j [g_{\theta}(x_{t_2})]_j \\
&=\sum_{i,j \in \theta \times \theta}  \frac{1}{|T^{(+)}|^2} \sum_{t_1,t_2 \in T^{(+)} \times T^{(+)}}  \frac{[g_{\theta}(x_{t_1})]_i \theta_i \theta_j [g_{\theta}(x_{t_2})]_j}{\|\hat{g_{\theta}}(x_{t_1})\| \|\hat{g}_{\theta}(x_{t_1})\|}  = \sum_{i,j \in \theta \times \theta} [\hat{\psi}^+_{M_{\theta}}]_{ij}\\
\end{align*}
\noindent where we denote the set of examples with label ``$+$'' as $T^{(+)}$, a subset of known examples $T$ and $M_{\theta}^{1/2} g_{\theta}(x)$ with $\hat{g_{\theta}}(x)$. Similar calculation can be applied if the labels are different thus 
\begin{align*}
\gamma_{M_{\theta}} &= \mathbf{E}_{x,y \sim P}[K_{M_{\theta}}(x,x') | l(x) = l(y)] - \mathbf{E}_{x,y \sim P}[K_{M_{\theta}}(x,x') | l(x) \neq l(y)] \\
& \approx  \sum_{i,j \in \theta \times \theta} [\hat{\psi}^{(+)}_{M_{\theta}}]_{ij} -  [\hat{\psi}^{(-)}_{M_{\theta}}]_{ij} =  \sum_{i,j \in \theta \times \theta} [\hat{\psi}_{M_{\theta}}]_{ij}. 
\end{align*}
\noindent Therefore if we consider multiple layers, where $[M_{\theta}]_{ij} = 0$ if $l(i)\neq l(j)$,  the \textit{similarity gap} can be approximated as 
\begin{align*}
\gamma_{M_{\theta}} \approx \sum_{i,j \in \theta \times \theta} \delta_{l(i),l(j)} [\hat{\psi}_{M_{\theta}}]_{ij} = \sum_{k \in {1,...,L}} \sum_{i,j \in \theta^k \times \theta^k}  [\hat{\psi}_{M_{\theta}}]_{ij}. 
\end{align*}
\noindent Note, with $|T| \rightarrow \infty$ the error of our approximation converges to zero with probability one, moreover this convergence is true for every element in $\hat{\psi}_{M_{\theta}}$. An additional consequence that we can partition the elements of the \textit{similarity gap} into disjoint sets and examine each independently. Observe, the elements of $\hat{\psi} \in \mathbb{R}$ therefore let us define for every pair their ``importance'' in the gap as $\textit{imp}_{i,j} = \max\{0,[\hat{\psi}_{M_{\theta}}]_{ij}\}$. By setting the elements in the metric with low or negative importance we can define a \textit{sparse block-diagonal} similarity with an additional step, normalization. However our new metric can have the same dimensions as the original block-diagonal. Thus we can argue that according to $\textit{imp}_i= \sum_{j} \textit{imp}_{ij}, \forall i \in \theta$ we can select the most important parameters and delete rows and columns associated with less important parameters and form the \textit{elementwise block-diagonal} metric. 
 
So far we assumed arbitrary input and do not take advantage of the gradient graph of the network. To show a case when the normalized sparse block-diagonal similarity has better \textit{similarity gap} than the original normalized block-diagonal similarity we will assume that our input is a subgaussian random vector \cite{lattimore2020bandit} with zero mean and variance one, e.g. the input is element-wise standard normalized. Additionally, observe that in ReLU networks the partial gradients can be expressed as $g_{\theta}(x) = S_{\theta}(x) x$ where $S_{\theta}(x)_{i,j} = \frac{\partial^{2}f_{\theta}(x)}{\partial \theta \partial x}\rvert_{\theta, x}$ a $d \times |\theta|$ sized matrix and inside an \textit{activation region} $S_{\theta}(x) = S_{\theta}(A(x))$ is identical for each $x \in A$ thus $g_{\theta}(x) = S_{\theta}(A(x)) x$. Due the complexity of the proof and the page limit we only mention that, following Theorem 2.1 in \cite{hsu2012tail}, the norm of the elementwise sparse vector for points in the activation region $A$ is concentrated as for all $\delta>0$
\begin{align*}
P(\|S^*_{\theta}(A(x)) x\|_2^2 > Tr[S^*_{\theta}(A(x)) S^*_{\theta}(A(x))^T] (1+4\delta); x \in A) \leq e^{-\delta}
\end{align*}
\noindent where $S^*_{\theta}(A(x))$ is the same as $S_{\theta}(A(x))$ without the removed rows and columns and therefore the \textit{similarity gap} in case of the normalized elementwise sparse block-diagonal similarity is related to the ratio $\frac{Tr[S_{\theta}(A(x)) S_{\theta}(A(x))^T]}{Tr[S^*_{\theta}(A(x)) S^*_{\theta}(A(x))^T]}$.

\begin{figure}[ht]
\centering
\includegraphics[width=0.48\textwidth]{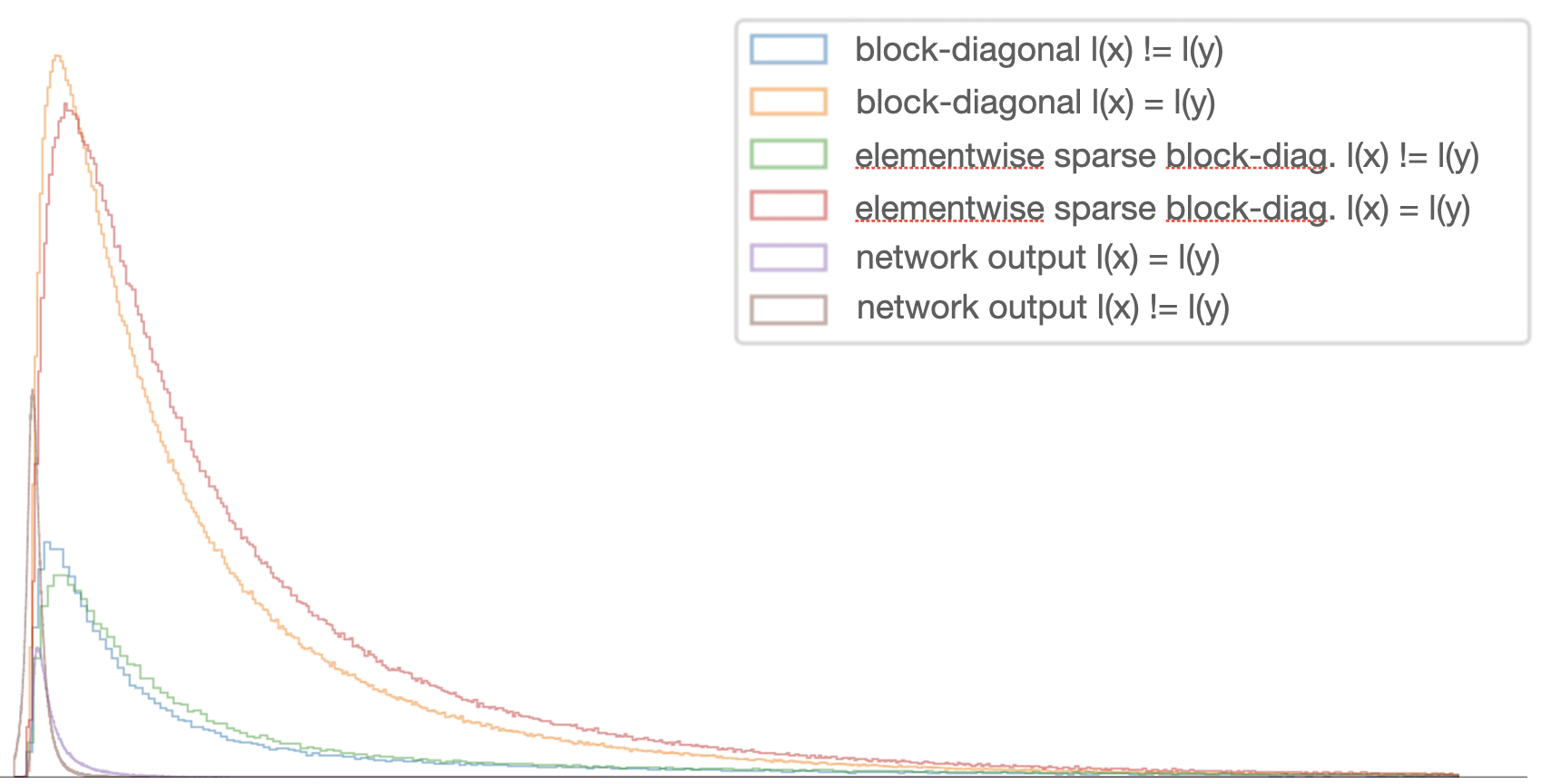}
\caption{Distribution of the network output, the \textit{elementwise sparse block-diagonal} and the \textit{block-diagonal} similarity values.}
\label{fig}
\end{figure}

We experimented on the CIFAR-10 dataset \cite{krizhevsky2009learning} with a simple feed-forward network with five layers. We ranked the parameters per layer according the \textit{elementwise block-diagonal} metric and deleted the parameters with low ``importance''. Results in Fig.~\ref{fig} indicate that the gap can be increased but further, more detailed experimentation needed to understand how the gap is actually increasing. 

\section{Conclusions}

In this paper we defined similarity functions over ReLU networks based on the gradient structure and investigated the \textit{similarity gap}. Furthermore, we introduced a measure to rank the parameter pairs in the network according to their importance in the \textit{similarity gap}. In the future we plan to extend our work to other network structures as our findings were limited to feed-forward fully connected ReLU networks.

\begin{footnotesize}

\bibliographystyle{unsrt}
\bibliography{rad_bib,tangent}

\end{footnotesize}


\end{document}